# Reasoning, Metareasoning, and Mathematical Truth: Studies of Theorem Proving under Limited Resources

Eric Horvitz
Decision Theory Group
Microsoft Research
Redmond, WA 98052
horvitz@microsoft.com

Adrian Klein*
Center for the Study of Language and Information
Stanford University
Stanford, CA 94305
klein@proof.stanford.edu

## Abstract

In earlier work, we introduced flexible inference and decision-theoretic metareasoning to address the intractability of normative inference. Here, rather than pursuing the task of computing beliefs and actions with decision models composed of distinctions about uncertain events, we examine methods for inferring beliefs about mathematical truth *before* an automated theorem prover completes a proof. We employ a Bayesian analysis to update belief in truth, given theorem-proving progress, and show how decision-theoretic methods can be used to determine the value of continuing to deliberate versus taking immediate action in time-critical situations.

## 1 INTRODUCTION

Theorem proving is frequently perceived as a logical, deterministic endeavor. However, uncertainty often plays a critical role in theorem proving and other mathematical pursuits. The mathematician George Polya emphasized the importance of plausible reasoning for guiding the intuitions and effort of mathematicians. In particular, he discussed the important role of analogy and uncertainty during mathematical theorem proving [Polya, 1954a, Polya, 1954b]. Conjectures about mathematical truth often draw upon existing proofs of related concepts and the results of inductive reasoning. As an example, logical knowledge and inductive evidence strongly bolstered the beliefs of many mathematicians about the truth of Fermat's last theorem before the recently developed proof of the theorem became available [van der Poorten, 1995b]. van der Poorten has commented on the importance of induction and intuition in belief about mathematical truth with regard to conjectures about the quantity of regular primes [van der Poorten, 1995a]:

* Current address: Microsoft, Redmond, WA 98052

But even now, we still cannot *prove* that there are infinitely many regular primes. Sure we 'know' from experiment—extensive computation, and from heuristics—the feeling in our stomachs...

We will formalize aspects of plausible reasoning in mathematics with an analysis of uncertain beliefs during theorem proving under limited computational resources. Specifically, we present a Bayesian analysis of the truth of a propositional claim, based on information about the size of the search space and progress towards a goal during theorem-proving. We show how decision-theoretic reasoning and metareasoning can be used to make decisions about the best actions to take, and about the duration of deliberation before taking action with a partial result. We characterize the expected value of computation for a theorem prover, allowing decision-making systems to determine how long to perform inference in a time-critical setting. Before delving into theorem proving, we will review briefly earlier work on rational action under limited resources, that has taken advantage of the use of flexible inference procedures and decision-theoretic metareasoning [Horvitz, 1987, Horvitz et al., 1989]. The previous work targeted problems with the ideal control of inference and representation with belief networks and influence diagrams in time-critical contexts. This work extends decision-theoretic reasoning and metareasoning under limited resources by highlighting the salience of uncertainty in the realm of theorem proving.

## 2 DECISION-THEORETIC METAREASONING

We are typically uncertain about the outcome of complex computational processes. Nevertheless, we can often characterize the results of computation with abstract descriptions of expected output. For example, we may be able to define attributes of quality and predict how these various dimensions will be refined with the allocation of resources. In reasoning about the quantity of time and memory to allocate to the refine-ment of partial results, we must often explicitly consider the uncertainty about the results of computation. The expected value of computation (EVC) is the difference between the expected utility of employing the current result immediately and the expected utility of the result obtained after allocating additional computational resources [Horvitz, 1988, Russell, 1990]. The net EVC (NEVC) is the difference in expected utility, including the cost associated with delayed action.

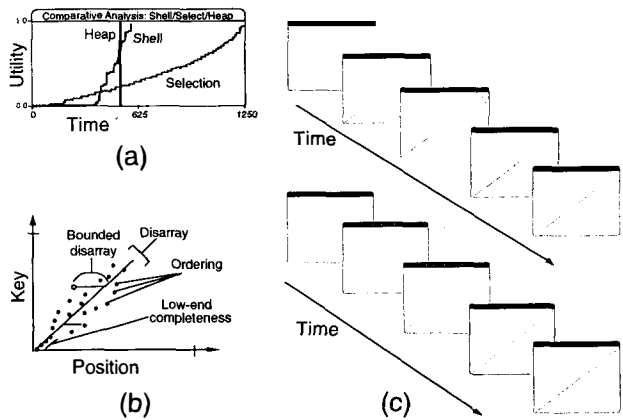

Figure 1: Early demonstrations of flexible computation with sorting algorithms. (a) Comparison of the utility of partial results generated by shellsort, selection sort, and less flexible heapsort algorithms. (b) Graph highlighting multiple attributes of value in a partial result of sorting. (c) Graphical representation of partial results during refinement of shellsort (lower sequence) and selection sort generated by the Protos/Algo system.

*Flexible computation* procedures are algorithms that provide continuous refinement of attributes of a computed result, given increasing allocation of one or more classes of computational resources (see [Dean and Wellman, 1991] and [Horvitz, 1987] for reviews of temporally flexible algorithms). Resources can include time and memory. Another desirable property of flexible methods for computing under varying resources is *convergence* on the ideal result with some finite or infinite quantity computational resources. This property allows reasoning systems to converge on ideal results with increasing amounts of computation. Convergence can be important in making arguments about *bounded optimality*—optimization of problem solving and action given the expected challenges and costs of computation (for general discussions of bounded optimality and rationality, see [Horvitz, 1987], [Doyle, 1990], and [Russell et al., 1993]). Dean and Boddy introduced the term *anytime algorithms* to describe flexible computation for incrementally refining the quality of plans generated by a planner [Dean and Boddy, 1988].

In brief, a flexible algorithm transforms a current problem instance $I$ into a partial result $\pi(I)$, consuming computational resource $r$. We often can decompose the comprehensive utility of a partial result into the inference-related costs, $u_i$, and the object-level value, $u_o$, of the result. The NEVC of allocating a quantity of resource $r$ to a flexible strategy $S_i$ to refine a problem instance $I$ is,

$$\mathrm{NEVC}(S_i, I, r) = \int_{\pi(I)} u_o(\pi(I)) \times p(\pi(I) \mid S_i, I, r) \\ - u_o(I) - u_i(r) \qquad (1)$$

The current problem instance $I$ may be a partial result $\pi^o(I)$, computed earlier by a flexible procedure with some prior allocation of resource. We may be uncertain about the multiple attributes or dimensions of value in a result, and about the utility functions $u_o$ and $u_i$ used to map object-level utility to attributes of partial results, and disutility to allocated resources. To handle such cases, we can generalize Equation 1 by summing over the uncertainty associated with utility assignments and uncertainty over different attributes of partial results.

We can employ EVC analyses to make decisions about the best flexible algorithms to apply and the length of time to apply the algorithms [Horvitz, 1988, Boddy and Dean, 1989]. Methods for handling such prototypical cost contexts as deadlines, uncertain deadlines, and general urgency are described in [Horvitz, 1988]. In general, real-time inference and information gathering for metareasoning must be tractable or compiled into tractable procedures through offline analysis [Horvitz, 1989]. Researchers typically have made greedy, myopic assumptions to keep metareasoning about ideal deliberation plans tractable.

## 3 RELATED WORK ON BELIEFS AND LIMITED RESOURCES

We shall focus on methods for controlling deliberation in a theorem prover by analyzing the dynamic changes in beliefs about truth with deliberation. In related work on computation under bounded resources, decision-theoretic methods have been used to reason about probability and action, and to determine ideal deliberation [Horvitz et al., 1989]. Research on the Protos project at Stanford focused specifically on the control of probabilistic and decision-theoretic inference in belief networks and influence diagrams via the use of tractable, approximate decision-theoretic metareasoning. The Protos system has served as an example of *reflective decision-analytic* reasoning—decision-theoretic reasoning that includes the costs of reasoning in computing optimal actions [Good, 1952, Horvitz, 1990]. In this work, a partial result is a second-order probability distribution over future prob-abilities or utilities. Protos applies flexible probabilistic inference to tighten the upper and lower bounds of probabilities that are required in time-critical decision problems. Sample problems were drawn from emergency and intensive-care medicine [Horvitz and Rutledge, 1991]. The system continues to monitor the EVC of decision-theoretic inference based on the state of the decision problem and meta-knowledge, obtained via analysis of performance of the inference algorithm, about the expected refinement of partial results with additional computation. When the net EVC becomes nonpositive, deliberation ceases and action is taken in the world. Figure 2 shows instrumentation output by Protos, including convergence of bounds on a probability with computation, time-dependent changes in utilities and decision thresholds, and graphical explanation of the best decision.

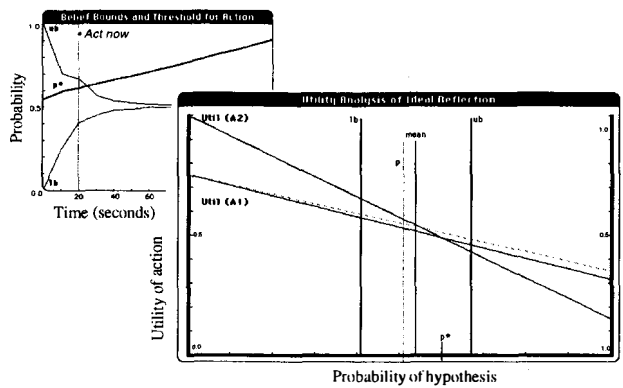

Figure 2: Decision-theoretic metareasoning about flexible probabilistic inference for a time-critical medical decision problem. Graph on left displays the convergence of bounds on probability with computation, the changing decision threshold, and ideal time to act. Larger graph shows Protos' explanation of partial result at time of action, including bounds on a critical probability and losses based in delay.

Other research on the development of flexible methods for computing beliefs under varying resources includes work on anytime probabilistic logic [Frisch and Haddawy, 1995] and on flexible decision making with a probabilistic database [Pittarelli, 1994]. In these analyses, incremental-refinement methods are presented that allow the number of sentences about beliefs and action to be manipulated, inducing tradeoffs in precision and computation that provide opportunities for control.

## 4  THEOREM-PROVING METHODOLOGIES

Let us now delve into the realm of theorem proving under bounded resources. We shall apply methods analogous to the previous work on beliefs and actions under bounded resources; specifically, we shall identify flexible procedures and partial results for refining belief in mathematical truth and for controlling theorem-proving deliberation. In the case of theorem proving, we are interested in whether a formula or a group of formulae (the conclusion) is implied by another set of formulae (the premises). Automated theorem provers compute the answer to such questions.

An infinite number of theorems can be deduced from the premises and logical tautologies. Thus, it is impractical to generate the conclusion from the premises through repeated application of valid rules of inference. An effective strategy is to show that a contradiction is implied by the conjunction of the premises, $P$, with the negation of the conclusion, $\neg C$. If a contradiction is found, then the conclusion must follow from the premises. If the premises themselves are inconsistent, then any set of formulae will follow.

Generative approaches to theorem proving, such as resolution refutation, do not provide indications how close the algorithm is to termination. The size of the theorem set at any given stage does not give us useful information about the likelihood that the set itself is consistent. However, other methods provide better handles into solution progress. We have studied a class of propositional theorem provers which implement the matrix method [Bibel, 1987]. The *matrix method* of theorem proving structures the task of checking the consistency of a set of statements as a search through a space of truth assignments. The matrix method has properties that make possible the gathering of information about the expected distance to a proof's completion.

The matrix method works to prove inconsistency by searching for a truth assignment which satisfies the clauses in $(P \wedge \neg C)$. If the search for a truth assignment fails, then the conclusion is entailed by the premises. The process is as follows: First, the conjunction of all formulae in the set $(P \wedge \neg C)$ is translated into conjunctive normal form. This results in a set of disjunctions or clauses, $C_1, C_2, \ldots, C_n$, which are implicitly conjoined. A *path $x$*, through this set of clauses, is a set of ground literals, $L_1, \ldots, L_n$, where each $L_i$ occurs in the disjunction $C_i$. Thus, $L_i, 1 \leq i \leq n$, is either a proposition $P$ or its negation. A path is said to be *open* iff $L_i \neq \neg L_j$, for all $1 \leq i, j \leq n$. A path is *closed* iff $L_i = \neg L_j$ for some $i, j$ $1 \leq i, j \leq n$. A truth assignment $v$ can satisfy all literals in a path $x$ iff $x$ is open. Since the set of formulae $(P \wedge \neg C)$ is logically equivalent to the set of clauses $C_1, C_2, \ldots, C_n$, a truth assignment exists that satisfies $(P \wedge \neg C)$ iff an open path exists through the clauses.

The search for an open path in the matrix method can be implemented as a depth-first search through a tree in which the literals of the first disjunct are the children of the root node, and all nodes at any given level have the literals of the next disjunct as their children. The search down any given subpath terminates whenever a contradiction is found (i.e., when a syntactic comparison shows that both a proposition and its negation occur on the current subpath), and all paths extending from this subpath are also discarded, since they will necessarily be contradictory as well. The conclusion must follow from the premises if no truth assignment can be found, equivalent to closing all paths. The matrix-method theorem-proving procedure must consider a search space of paths bounded by the number of literals raised to the power of the number of clauses. For example, in a matrix with $n$ clauses, where each is a disjunction of $m$ literals, there are $m^n$ possible paths through the matrix.

## 5  PROBABILITY AND PROOF

What is the link between theorem proving and probability? Theorem proving with the matrix method involves checking whether each of the $2^k$ possible truth assignments for the set of literals appearing in the problem satisfies the members of $(P \wedge \neg C)$. With matrix theorem proving, the search space that is traversed in practice is not isomorphic to the space of possible valuations or models. However, there is a correspondence between the syntactic search for an open path and the semantic search through a space of possible propositional truth assignments. Each subpath $x$ through clauses $C_1, C_2, \ldots, C_n$, corresponds to the intersection of the class of truth assignments $v$ in which $v(P) = true$ if $L_i = P$ for some $1 \leq i \leq n$, with the class of truth assignments $v'$ in which $v'(P) = false$ if $L_i = \neg P$ for some $1 \leq i \leq n$. Since the correspondence between syntactic paths and possible truth assignments is one-to-many, a search through the former space will generally be completed more quickly than a truth-table style examination of the latter. Also, within a particular matrix, each path of a given length will contain roughly the same number of different literals, and so will correspond to approximately the same number of truth assignments. Thus, the interpretation of the portion of the search space explored can be extended from the semantic space of truth assignments to the syntactic domain of paths explored in the matrix method.

During matrix theorem proving, if a contradiction is produced by adding a literal $L_i$ from a clause numbered $n_j$ to the current subpath, when that already contains an occurrence of $\neg L_i$, the number of paths of the total initial search space $m^n$ that have already been searched is incremented by an additional $m^{(n-n_j)}$ paths; this is the number of complete paths containing this subpath. By recording the occurrences of such path closings, we can track the frequency of closings as a function of the portion of the total search space visited and record probabilistic information about the search space explored before an open path is discovered by the reasoner. We use such search information to compute the probability of truth before a proof is completed.

Assume that $w$ represents the metatheoretic claim that a conclusion follows from the cited premises. We wish to compute the probability that the matrix method will determine the truth of $w$, given information that some portion of total search space has been explored without discovery of an open path. We condition our analysis on the absence of a logical proof. We will use $S$ to indicate the portion of the total search space that has been explored without finding a proof of $w$. We can compute the probability of $w$ by employing Bayes' rule to relate the probability that $w$ is true to evidence about the progress of search and the prior probability of truth,

$$p(w|S,\xi) = \frac{p(S|w,\xi)p(w|\xi)}{p(S|w,\xi)p(w|\xi) + p(S|\neg w,\xi)p(\neg w|\xi)} \quad (2)$$

where $p(w|S,\xi)$ is the likelihood of the truth of the metatheoretic claim $w$, given that $S$ of the search space has been explored without discovering an open path, and background (implicit) information about the situation $\xi$. The term $p(S|w,\xi)$ is the probability of $S$ given $w$ is true, $p(S|\neg w,\xi)$ is the probability of $S$ given $w$ is false, and $p(w|\xi)$ is the prior probability that $w$ is true. The prior probability of $w$ is based on experience with a set of queries to a theorem prover, conditioned on such information as the source, size, and structure of the input. The probability that $w$ is false, $p(\neg w|\xi)$, is simply the complement of the prior probability of $w$, $1 - p(\neg w|x)$.

We can simplify Equation 2 by noting that, when $w$ is true, no open paths can be found, and the theorem prover will search the entire space before halting. Thus for all fractions of the search less than one, $p(S|w,\xi) = 1$. Now, we can express $p(w|S,\xi)$ in terms of $p(w|\xi)$ and $p(S|\neg w,\xi)$,

$$p(w|S,\xi) = \frac{p(w|\xi)}{p(w|\xi) + p(S|\neg w,\xi)[1 - p(w|\xi)]} \quad (3)$$

Thus, we can compute the truth of a proposition before a theorem prover halts if we know the prior probability of the truth of $w$ and the probability that search will explore increasing portions of the total search space without finding a proof that $w$ is false, given that $w$ is indeed false.

What can we say about the expected form of the probability distribution $p(S|\neg w,\xi)$? In the absence of problem-specific information, we assert event equivalence regarding the likelihood that each possible truth assignment for the set of literals appearing in the problem will satisfy the members of $(P \wedge \neg C)$. Event equivalence implies that each of the $2^k$ possible truth assignments is equally likely to be consistent with $\neg w$.

Assume that each path in the search space is independent of other paths and that there are $\mathcal{O}$ open paths in the matrix. The probability of not finding a proof of $\neg w$ after searching portion $s$ of the total search space is,

$$p(S = s|\neg w, \xi) \approx \prod_{i=0}^{sm^n - 1} 1 - \frac{\mathcal{O}}{m^n - i} \qquad (4)$$

where $m^n$ is the total number of paths in the matrix. In this paper, we shall not demonstrate the explicit use of such information as the expected number of open paths $\mathcal{O}$. Rather, we show how data about $p(S|\neg w, \xi)$ can be employed directly. However, in the general case, we can seek from data the probability distribution over the number of open paths as a function of such distinctions as the size of the space of truth assignments, conditioned on the nature and source of the problem instance. Factors, including the number of clauses, clause length, and size of alphabet can contribute to the number of open paths. We can account for uncertainty in the number of open paths by modifying Equation 4 to consider a probability distribution, $p(\mathcal{O}|I_1...I_n, \xi)$, representing information about the number of open paths conditioned on problem-instance attributes, $I_i$.

We know that paths in a matrix are not independent; because of the branching structure of the paths in the matrix, paths can share a large subset of ancestors. Given a large number of paths, however, we can pose an argument for minimal dependency by substituting *groups* of paths for literals for single paths. We group together those paths that have a large percentage of literals in common, and assume, without additional information, that each of these weakly dependent sets of paths have an equal chance of containing an open path.

## 6   EMPIRICAL STUDIES

We ran a large number of experiments and collected data on the relevance of the portion of search space explored without finding a proof to the likelihood that a proposition is true. Specifically, we collected information on the prior probability of truth of $w$ and the probability that a search would progress to portions of the search space given that $w$ is false, as required for determining the truth of $w$ with Equation 3.

### 6.1   EXPERIMENTS

In the experiments, we generated a large set of propositional clauses as inputs to the theorem prover and iterated the matrix algorithm over each. We restricted the inputs to a fixed number of clauses and fix the numbers of literals in each disjunction. The occurrences of path closings were recorded as a function of the proportion of the search space explored and the fraction of the total search space explored before an open path was discovered by the reasoner. We ran a large number of cases and used these multiple samples to determine the probability that a counterexample to the claim of entailment would be found with further computation as a function of the portion of the search space visited.

The propositional matrices for the studies were generated in a straightforward fashion: Each clause is restricted to the same number of literals, and each literal within a clause is generated randomly. A propositional symbol is selected from an alphabet of specified size, with the selection procedure being repeated until a symbol is found that is not already present in the current clause. This symbol is then negated with a 0.5 probability. The use of such randomly generated propositional clauses as a testbed for performance-enhancing heuristics has been investigated by [Mitchell et al., 1992] and [Selman et al., 1992]. For a fixed number of literals per clause, the ratio of the number of clauses $m$ comprising the problem instance to the size $a$, of the propositional alphabet employed is influential in determining the difficulty that problem instances will pose to automated theorem-proving routines.

If there are a large number of clauses in relation to the number of propositions represented in the matrix, the matrix will tend to be trivially unsatisfiable; contradictions within the matrix are so plentiful that very little processing work need be done to close all of its paths. Alternatively, if there are too few clauses, contradictions will be so scarce as to make the location of an open path within the matrix a simple matter. In a study of varying difficulty of satisfiability problems, Mitchell et al. identify the settings likely to produce the most challenging problem instances as those which tend to generate an equal number of satisfiable and unsatisfiable collections of clauses [Mitchell et al., 1992]. In our trials, the fraction of unsatisfiable matrices generated ranged between 0.3-0.6.

### 6.2   RESULTS: THEOREM-PROVING PROFILES

Data on theorem-proving behavior was collected by running the matrix method on 750 randomly generated propositional matrices, each containing 20 clauses composed of 3 literals apiece. The literals were generated from a 4-symbol propositional alphabet. We found that thirty percent of the matrices tested contained no open paths whatsoever; thus, the prior probability of $p(w|\xi)$ for matrices generated for the study is 0.3. Figure 3 shows the probability of finding an open path as a function of the portion of search space visited for cases where $w$ was found to be false. The probability of not finding a path as a function of the space searched, $p(S|\neg w, \xi)$, is just the complement of the plotted values. As described earlier, a cumulative probability curve with a binomial structure would suggest that the paths in the matrix, or suitable weakly dependent groups of paths, are equally likely to be open.

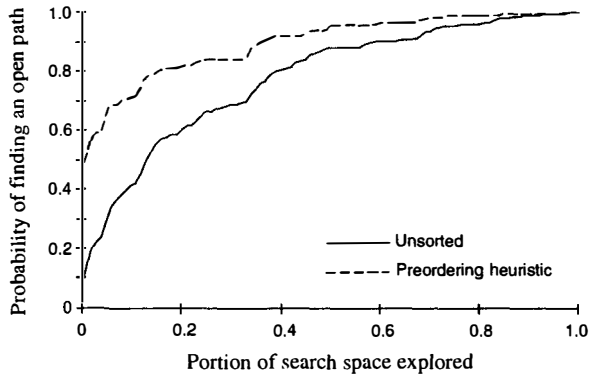

Figure 3: Experimental results. Probability of finding an open path, given $\neg w$, plotted as function of the portion of the search space explored (solid line). Results are based on 750 randomly generated propositional matrices, each containing 20 clauses that are composed of 3 literals each. Broken line displays results for identical set of instances after application of a presort heuristic.

We can use the data gathered in these experiments to compute the probability of $w$ before a proof is completed. As an example, assume that the theorem prover has proceeded through forty percent of the search space. The data shows that $p(S|\neg w, \xi) = 0.2$ when $s = 0.4$. Substituting this, and the prior value of truth, 0.3, into Equation 3 yields a probability of truth given the search, $p(w|S, \xi) = 0.68$.

### 6.3 ANALYSIS OF HEURISTICS

Heuristics have played a central role in research on automated theorem proving. We explored how we could use the Bayesian framework to probe the effectiveness of heuristics. We investigated the value and probabilistic implications of applying a preording heuristic that reorders the literals within a clause based on the results of the inspection of the literals in the matrix. A detailed discussion of the heuristic is described in [Klein and Horvitz, 1994].

We performed experiments with a preording heuristic. The broken line graphed in Figure 3 shows the probability of an open path being found for the sorted case, for the same set of cases. The curves for the sorted and unsorted cases show significant differences. As in the example for the unordered case, let us assume that the search has proceeded through 0.4 of the search space. The data indicates that for the theorem prover using the preorder heuristic $p(S|w, \xi) = 0.08$ when $S = 0.4$. Substituting this probability into Equation 3 yields a probability of truth given the search, $p(w|S, \xi) = 0.84$, in contrast to 0.68 for the unordered situation. In this case, the theorem prover employing the preorder heuristic and associated data would provide stronger belief in the truth of $w$, given an equivalent amount of search.

## 7 ACTION BEFORE PROOF

We now move from the realm of belief about truth to action in the world. Assume that the expected value of an agent's action depends on the truth of a formula which we can prove via the matrix method. Making inferences about the probability that an open path exists allows an agent using a propositional logic knowledge base to take action based on the partial results of incomplete theorem-proving, rather than being forced to wait until the termination of the logical analysis. As in other applications of flexible computation methods, we move from a traditional all-or-nothing analysis to one considering a spectrum of partial results.

### 7.1 IMMEDIATE ACTION AT A DEADLINE

An agent employing a theorem prover should take actions that maximize its expected utility. To compute the expected value of different actions $A_i$ in terms of the likelihood of the truth of one or more propositional formulae, we must consider the utilities of outcomes $u(A_i, w_j)$, $u(A_i, \neg w_j)$, for all actions $A_i$ and formulae $w_j$, and select the action that maximizes the expected utility (EU). Let us assume that the EU of taking action $A_i$ depends on the truth of multiple formulae, $w_j$, and that the formulae are mutually exclusive. In this case, the EU of taking action $A_i$ is

$$\text{EU}(A_i) = \sum_j p(w_j|S, \xi) u(A_i, w_j) \quad (5)$$

For cases where the best action is determined by belief in the truth of a single formula $w$, the best action, $A^*$, is,

$$A^* = \arg\max_A [p(w|S, \xi)(u(A_i, w) - u(A_i, \neg w)) + u(A_i, \neg w)] \quad (6)$$

For situations where there are only two actions under consideration (e.g., $A_1$=FIGHT, $A_2$=FLIGHT), we can summarize the best policy for action by considering the relationship of $p(w|S, \xi)$ to a *threshold probability*, $p^*$, the probability of truth in formula $w$ where the two actions have the same expected utility,

$$p^* = \frac{u(A_2, \neg w) - u(A_1, \neg w)}{u(A_2, \neg w) - u(A_1, \neg w) + u(A_1, w) - u(A_2, w)} \quad (7)$$

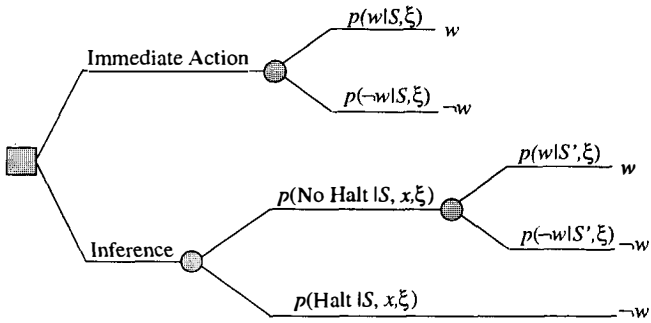

Figure 4: Decision making about immediate action versus continuing to deliberate with a theorem prover. To determine the value of deliberation, we compare the expected utility of acting immediately with the expected value of the lottery defined by a commitment to continue to explore $x$ additional paths, unless a proof is found.

In such binary decision problems, action $A_1$ is the best course of action if $p(w|S, \xi)$ is greater than $p^*$; if it is less than the threshold probability, action $A_2$ dominates.

## 7.2 IMMEDIATE ACTION VERSUS INFERENCE

We now introduce the expected value of computation for logical inference with the matrix method, $\text{NEVC}_l$. This measure allows us to consider the tradeoff between the cost associated with delayed action and the promise of making a better decision with additional computation. Assume we have already searched portion $S$ of the search space $s$ without discovering a proof. We wish to determine the value of searching an additional $j$ paths.

Figure 4 displays a decision tree for the problem of acting immediately versus delaying for additional deliberation. If a theorem prover is allowed to investigate $x$ additional of paths in the matrix method, two outcomes are possible: (1) the system will find a proof of $\neg w$ and will halt, or (2) the system will not halt. If the theorem prover does not halt, we update S to S', and revise our belief in the truth of $w$, $p(w|S', \xi)$, with Equation 3.

Unfortunately, we may have to pay a penalty for the delay associated with search of additional paths. In time-critical situations, the utility of one or more outcomes is dependent on the length of delay before action is taken. We can represent time-dependent utilities by extending the representation of the utility of an outcome, employed in Equation 6, to include changes in the utility associated with delays $t$ incurred before action, $u(A_i, w_j, t)$.

We shall consider the expected utility of exploring an additional $j$ paths without discovering a proof. We use $p(w|S, j, \xi)$ to refer to the probability of $w$ given that we search another $j$ paths without finding a proof, after previously searching a portion $S$ of the space. We use $t(j)$ to refer to the amount of time required to search $j$ paths. We first consider the value of acting immediately after searching the $j$ paths without finding a proof. We use $U(S, j)$ to refer to the expected utility of executing the best action after searching $j$ additional paths without finding a proof. If we act immediately after searching $j$ paths without finding a proof, the expected utility of the best action will be

$$U(S,j) = \\ \max_A [p(w|S, j, \xi)(u[A_i, w, t(j)] - u[A_i, \neg w, t(j)]) \\ + u(A_i, \neg w, t(j))] \qquad (8)$$

Let us first consider the $\text{NEVC}_l$ for the decision to continue to perform logical inference for a single additional path. We must take into consideration the probability that the theorem prover will find the next path to be open and will halt, and the case where it does not halt. We use $p(H|S, j, \xi)$, to refer to the probability that the system will halt on the $j$th additional path explored, concluding then that $w$ is false. The $\text{NEVC}_l$ for a single path is,

$$\text{NEVC}_l(S, 1) = \\ p(H|S, 1, \xi) \max_A u[A_i, \neg w, t(1)] \\ + [1 - p(H|S, 1, \xi)]U(S, 1) - U(S, 0) \qquad (9)$$

where $U(S, 0)$ refers to the expected utility of immediate action instead of undertaking additional search.

Given general temporal cost functions, a single-step analysis can fail to identify the possibility that there will be positive value in computing for longer periods of time. Thus, it may be useful to consider the $\text{NEVC}_l$ for searching multiple paths. A formulation of $\text{NEVC}_l$ for arbitrary numbers of future paths $x$, within the remaining search space, considers the probabilities that the system will halt at different times before all $x$ remaining paths have been explored,

$$\text{NEVC}_l(S, x) = \\ \sum_{j=1}^{x} p(H|S, j, \xi) \max_A u[A_i, \neg w, t(j)] \\ + \left[1 - \sum_{j=1}^{x} p(H|S, j, \xi)\right] U(S, j) - U(S, 0) \quad (10)$$

For a binary decision problem, the maximizations indicated in the equation can be performed simply by checking to see if the probability $p(w|S, \xi)$ is greater or less than $p^*$.

We can acquire the probability distributions necessary to solve Equations 9 and 10 directly from the data collected about the performance of the theorem prover as described in Section 5. However, we can also employ probability models that explain the relationships seen in the data, as we described with Equation 4. In particular, we can approximate the probability of halting on the $j$th new path searched. When $w$ is true, the probability of halting on any of the $j$ additional branches of the search before completing the search, $p(H|w, S, j, \xi)$, is zero; when $w$ is true, the theorem prover will not halt until exhausting the entire space. Thus, we need only to consider the probability of halting for the case where $w$ is false,

$$p(H|S, j, \xi) = p(H|\neg w, S, j, \xi) p(\neg w|S, j, \xi) \quad (11)$$

If we make similar assertions of independence as those assumed in formulating the probability model in Equation 4, the first term of Equation 11 can be approximated as,

$$p(H|\neg w, S, j, \xi) \approx \left[\frac{\mathcal{O}}{l-(j+1)}\right] \prod_{i=0}^{j-2} 1 - \frac{\mathcal{O}}{l-i} \quad (12)$$

assuming we search $j$ additional paths, leaving $l$ paths of the total search space unexplored. The probability $p(\neg w|S, j, \xi)$ can be computed with Bayes' theorem as described in Equation 3.

## 8  OPPORTUNITIES IN FIRST-ORDER LOGIC

We have investigated theorem proving with a propositional language. Applications of theorem proving may require a more expressive first-order language (FOL). We can extend the matrix method to a domain of first-order clauses with ease. However, a first-order language introduces complications for the interpretation of the search space explored as the probability of truth. With FOL, it is difficult to deterministically analyze the size of the search space, and to determine or approximate the ratio of search space explored to remaining search space. Indeed, the size of the FOL search space is unstable as the space can continue to expand because of instantiation during the processing of a matrix. Difficulty with the a priori assessment of the size of the search space introduces difficulties with determining the fraction of the total search space explored, and with harnessing information about the progress through the space to compute the probability that a conclusion must follow from premises.

We believe the extension of decision-making methods to FOL will be a challenging and promising area research area. Several methods may be employed to extract useful information about the probability of truth in first-order logic. These include the use of a probabilistic analysis of the size of the search space, and portion of search space explored, conditioned on evidence about the problem instance and on information gleaned during preprocessing and search. The probabilistic and decision-theoretic analyses described in this paper can be extended with probability distributions over these quantities. Additional theoretical and empirical studies may reveal approaches to gaining access to such probability distributions over the size of the search space size.

## 9  SUMMARY

We reviewed related work on flexible computation and control of deliberation for computing beliefs and actions under bounded resources. We took a decision-making perspective on theorem proving, focusing on the use of theorem-proving activity to guide decisions about additional deliberation and about actions in the world. We hope that this work provides a valuable conceptual bridge between theorem proving under limited computational resources and probabilistic reasoning. We foresee research on probabilistic methods for tackling problems with the analogous analysis of FOL theorem proving. In particular, probabilistic methods may prove useful for inducing the expected size of a search space generated during FOL theorem proving, and for harnessing information about the progress of a theorem proving system through the search space with continuing computation.

### Acknowledgments

We thank Jack Breese, Johann Dekleer, David Heckerman, Paul Lehner, Sheila McIlraith, Olivier Raiman, and Dave Smith for useful comments and suggestions on this work.

We can acquire the probability distributions necessary to solve Equations 9 and 10 directly from the data collected about the performance of the theorem prover as described in Section 5. However, we can also employ probability models that explain the relationships seen in the data, as we described with Equation 4. In particular, we can approximate the probability of halting on the $j$th new path searched. When $w$ is true, the probability of halting on any of the $j$ additional branches of the search before completing the search, $p(H|w, S, j, \xi)$, is zero; when $w$ is true, the theorem prover will not halt until exhausting the entire space. Thus, we need only to consider the probability of halting for the case where $w$ is false,

$$p(H|S, j, \xi) = p(H|\neg w, S, j, \xi) p(\neg w|S, j, \xi) \quad (11)$$

If we make similar assertions of independence as those assumed in formulating the probability model in Equation 4, the first term of Equation 11 can be approximated as,

$$p(H|\neg w, S, j, \xi) \approx \left[\frac{\mathcal{O}}{l-(j+1)}\right] \prod_{i=0}^{j-2} 1 - \frac{\mathcal{O}}{l-i} \quad (12)$$

assuming we search $j$ additional paths, leaving $l$ paths of the total search space unexplored. The probability $p(\neg w|S, j, \xi)$ can be computed with Bayes' theorem as described in Equation 3.

## 8  OPPORTUNITIES IN FIRST-ORDER LOGIC

We have investigated theorem proving with a propositional language. Applications of theorem proving may require a more expressive first-order language (FOL). We can extend the matrix method to a domain of first-order clauses with ease. However, a first-order language introduces complications for the interpretation of the search space explored as the probability of truth. With FOL, it is difficult to deterministically analyze the size of the search space, and to determine or approximate the ratio of search space explored to remaining search space. Indeed, the size of the FOL search space is unstable as the space can continue to expand because of instantiation during the processing of a matrix. Difficulty with the a priori assessment of the size of the search space introduces difficulties with determining the fraction of the total search space explored, and with harnessing information about the progress through the space to compute the probability that a conclusion must follow from premises.

We believe the extension of decision-making methods to FOL will be a challenging and promising area research area. Several methods may be employed to extract useful information about the probability of truth in first-order logic. These include the use of a probabilistic analysis of the size of the search space, and portion of search space explored, conditioned on evidence about the problem instance and on information gleaned during preprocessing and search. The probabilistic and decision-theoretic analyses described in this paper can be extended with probability distributions over these quantities. Additional theoretical and empirical studies may reveal approaches to gaining access to such probability distributions over the size of the search space size.

## 9  SUMMARY

We reviewed related work on flexible computation and control of deliberation for computing beliefs and actions under bounded resources. We took a decision-making perspective on theorem proving, focusing on the use of theorem-proving activity to guide decisions about additional deliberation and about actions in the world. We hope that this work provides a valuable conceptual bridge between theorem proving under limited computational resources and probabilistic reasoning. We foresee research on probabilistic methods for tackling problems with the analogous analysis of FOL theorem proving. In particular, probabilistic methods may prove useful for inducing the expected size of a search space generated during FOL theorem proving, and for harnessing information about the progress of a theorem proving system through the search space with continuing computation.

### Acknowledgments

We thank Jack Breese, Johann Dekleer, David Heckerman, Paul Lehner, Sheila McIlraith, Olivier Raiman, and Dave Smith for useful comments and suggestions on this work.